\documentclass[a4paper,12pt]{elsarticle}
\usepackage{amsmath}
\usepackage{amsfonts}
\usepackage{amssymb}
\usepackage{graphicx}
\usepackage{bm}

\usepackage{floatrow}
\usepackage{subcaption}
\newfloatcommand{capbtabbox}{table}[][\FBwidth]
\biboptions{authoryear}
 
\begin{document}
\begin{frontmatter}
\title{When can we improve on sample average approximation for stochastic optimization?}
\author[label2]{Eddie Anderson}
\ead{edward.anderson@sydney.edu.au}
\author[label2]{Harrison Nguyen\corref{cor1}}
\ead{harrison.nguyen@sydney.edu.au}
\cortext[cor1]{Corresponding author}
\address[label2]{The University of Sydney, NSW, 2006, Australia}
\begin{abstract}
We explore the performance of sample average approximation in comparison with several other methods for stochastic optimization when there is information available on the underlying true probability distribution. The methods we evaluate are (a) bagging; (b) kernel smoothing; (c) maximum likelihood estimation (MLE); and (d) a Bayesian approach. We use two test sets, the first has a quadratic objective function allowing for very different types of interaction between the random component and the univariate decision variable. Here the sample average approximation is remarkably effective and only consistently outperformed by a Bayesian approach. The second test set is a portfolio optimization problem in which we use different covariance structures for a set of 5 stocks. Here bagging, MLE and a Bayesian approach all do well.     
\end{abstract}

\begin{keyword}
stochastic optimization \sep sample average approximation \sep maximum likelihood estimation \sep bagging \sep kernel smoothing
\end{keyword}
\end{frontmatter}
\section{Introduction}

We suppose that we have a sample of $N$ points $S=\{y_{1},y_{2},....y_{N}\}$ drawn from a random variable $Y$ with an unknown distribution and we wish to solve the stochastic optimization problem:
\[
\text{P: }\min_{x \in X}\mathbb{E}[c(x,y)],
\]
where $x$ is the decision variable and lies in $X$ and the expectation is taken over the unknown distribution for $Y$. Typically such problems arise when a decision needs to be taken now (e.g. investment) to optimize a future payoff that depends on a random variable (e.g. demand each day) where the only information on that random variable is a historical sequence (e.g. daily demand over the last year).

The standard approach is to use the sample average approximation (SAA), so that we solve the problem:
\[
\text{SAA: }\min_{x \in X}\frac{1}{N}\sum c(x,y_{i})
\]
in which we treat the sample as though it was the full distribution. This amounts to optimizing the objective within sample, and then applying the optimal decision out of sample. In fact an SAA approach is frequently used without identifying it as such, when researchers concentrate on the sample at hand and have no way to assess directly the quality of the solutions they achieve.

The SAA method is generally very effective and makes no assumptions at all about the underlying distribution for the random variable $Y$. However in practice we are likely to have at least some information about the distribution. Perhaps we know that the underlying density function is continuous, or smooth. Perhaps we know that the underlying distribution has a single mode. Or perhaps we know that there is a good approximation to the underlying distribution from a parameterized set of distributions. We are interested in the following question: when and how can we use information about the underlying unknown distribution to improve on the solution obtained by SAA?

If there is a large sample it is easy to see that the problem SAA will be close to the underlying problem P, and there are results in the literature which make this statement more precise. Specifically a result of \cite{shapiro2009lectures} shows convergence under relatively weak conditions (such as Lipschitz behaviour for the cost function in terms of $x$ and compactness of $X$). Thus with a large sample the room for improvement on SAA will be small and we will have a harder task to establish whether or not a given method is worthwhile applying. Thus we will concentrate on problems with small samples ($N\leq200$).

Some recent work (\cite{esfahani2018data}, \cite{gotoh2017calibration}) on distributionally robust optimization has shown that in some circumstances these methods give better out of sample performance than SAA. But we will not include these approaches in our comparisons.

In this paper we report on a series of numerical experiments that address the question of whether we can improve on SAA. The results will demonstrate how effective SAA is, since it is surprisingly hard to show improvements and these are usually modest. One reason for this is that the use of more sophisticated approaches gives rise to a greater danger of overfitting, where the characteristics of the particular random sample $S$ are fitted by a postulated density function in a precise way that would not be justified if proper allowance was made for the uncertainty within the sample.

We have experimented with a few different approaches but we believe that there are four which are worth discussing in detail as alternatives to straight SAA: (A) Bagging; (B) Kernel Smoothing; (C) Maximum Likelihood Estimation; and (D) a Bayesian approach. We summarize our main conclusions as follows. On our test problems, a Bayesian approach is very effective. The maximum likelihood approach has variable performance that depends on the problem structure and the number of parameters to be estimated. For portfolio optimization it performs well, but for one dimensional problems it does not offer significant improvement. For multivariate problems bagging is a technique that can improve results substantially and should be considered as a matter of course. However using a smoothing technique to estimate the underlying density does not improve performance.

In this type of numerical study the conclusions we arrive at are inevitably tentative. We hope that this work will act as a spur to increased understanding of this issue and the variety of methods that may be applied. We make use of techniques that are drawn from both the machine learning and stochastic optimization areas and there is value in comparing these different approaches within a common framework.

The paper is laid out as follows. In the next section we introduce the four alternatives to SAA that we consider. In section 3 we describe the computational experiments that we have carried out. Finally in section 4 we discuss the results of our experiments and conclude.

\section{Four Alternatives to Sample Average Approximation}

For each of the four alternatives we will consider the arguments that suggest the approach may be effective, as well as the information assumptions on the underlying distribution, i.e. the nature of the additional information that is available to the decision maker. This may be knowledge that the underlying density for the random variable is smooth, or, for a better informed decision maker, the parameterized set of distributions from which the sample is drawn.

\subsection{Bagging}

Bagging refers to the technique of bootstrap aggregation. The process is quite simple: we take a new sample of size $M$ from $S=\{y_{1},y_{2},...,y_{N}\}$ with replacement, to obtain $S^{(1)}=\{y_{1}^{(1)},y_{2}^{(1)},...,y_{M}^{(1)}\}$ and then solve the problem SAA for the sample $S^{(1)}$ to find a solution $x^{(1)}$. We repeat this process $B$ times, to generate samples $S^{(1)},S^{(2)},...,S^{(B)}$ with corresponding SAA solutions $x^{(1)},x^{(2)},...,x^{(B)}$ and we finally form the bagging solution by averaging, so $x_{bag}=(1/B)\sum_{j=1}^{B}x^{(j)}$. We can then test $x_{bag}$ out of sample and, as we will see, there may be an  improvement in comparison with the SAA solution $x_{SAA}(S)$ from the original sample $S$, even though $x_{bag}$ is based on exactly the same underlying set of $y$ values as $x_{SAA}(S)$. Often we take $M=N$,but the method can be applied with $M<N$ or even $M>N$. Also the description we have given assumes sampling with replacement (i.e. a bootstrap), but as observed by Buja and Stuetzle \cite{buja2006observations} we can also use a subsampling approach where we take $M<N$ (say $M=N/2$) and create the samples $S^{(1)},S^{(2)},...S^{(B)}$ by sampling without replacement.

\begin{figure}
    \centering
    \includegraphics[width=16cm]{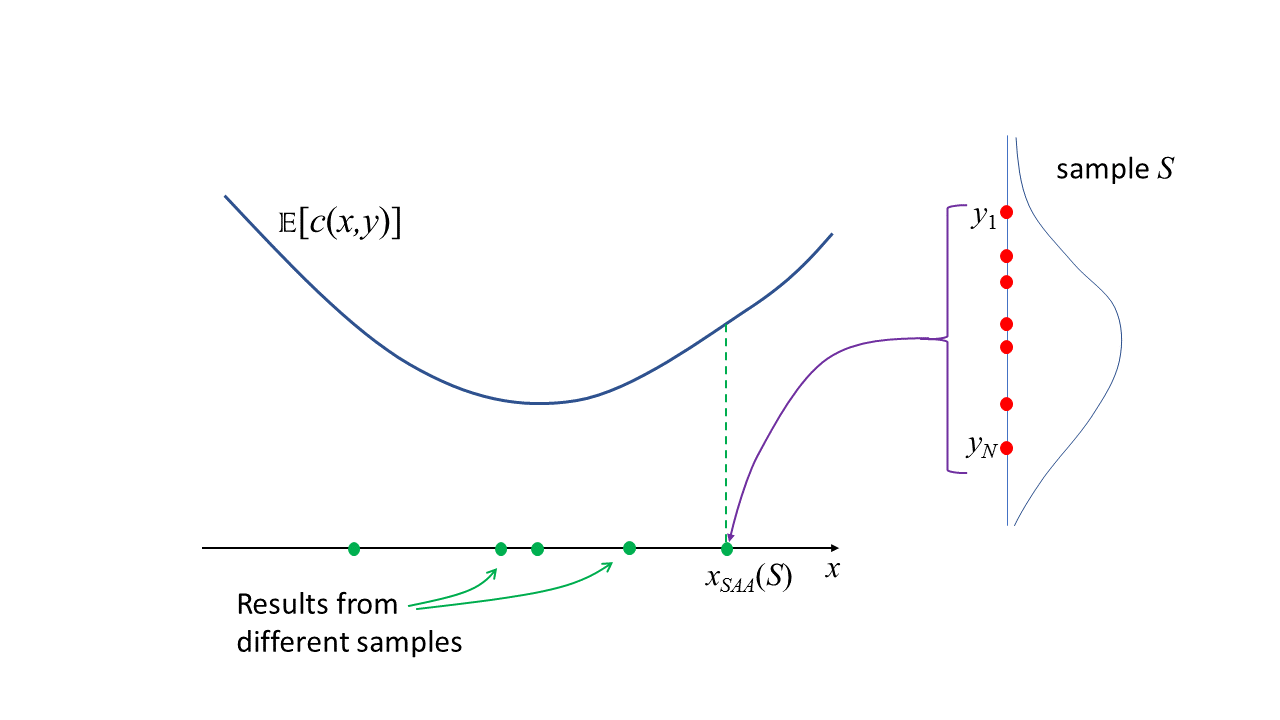}
    \caption{Diagram to illustrate why shrinkage is helpful.}
    \label{fig:shrinkage}
\end{figure}

Just like SAA, the bagging approach does not make any assumptions on the underlying distribution for $y$. The key argument that suggests we may be able to improve on a standard SAA approach is related to what is usually called shrinkage. This is illustrated in Figure \ref{fig:shrinkage} for a univariate decision variable $x$. The choices $x_{SAA}(S_{i})$ vary according to the choice of sample $S_{i}$ and we can expect to have some samples with the value $x_{SAA}(S_{i})$ being too high in comparison with the best possible value $x^{\ast}$ which minimizes the out of sample expectation $\mathbb{E}[c(x,y)]$. Equally there will be some samples where the value $x_{SAA}(S_{i})$ is too low. Thus we expect that averaging over the different $x_{SAA}(S_{i})$ values will produce a value that has better performance through a reduction in variance (though if we really had a number of different samples available it would be more natural simply to combine them into a larger sample and work directly with this). The bagging approach uses the same ideas but replaces the averaging over alternative samples with an average over different bootstrap samples all generated from $S$. This is known to reduce the sample induced variance for the estimator $x_{bag}$ in comparison with the estimator $x_{SAA}(S)$.

\subsection{Kernel Smoothing}

The problem we are faced with can be viewed as follows: we are given a sample from an unknown distribution and wish to estimate the underlying density function in order to carry out an optimization for the expectation with respect to this ``true'' density function. In most cases of interest in practice the underlying density will have some properties of smoothness, and it makes sense to use this information. The standard approach in statistics is kernel density smoothing, under which we estimate the density at a point $y$ as
\[
\widehat{f}(y)=\frac{1}{Nh}\sum_{i=1}^{N}K\left(  \frac{y-y_{i}}{h}\right)
\]
for some kernel function $K$ that integrates to $1$. We will choose a Gaussian kernel $K(y)=(1/\sqrt{2\pi})\exp(-y^{2}/2)$.

With kernel smoothing we will generate $\widehat{f}$ and then solve the problem
\[
\text{P}_{\text{smooth}}:\min_{x}\int c(x,y)\widehat{f}(y)dy.
\]
In our computational experiments this is done by taking a very large number of samples from the smoothed density $\widehat{f}$ and applying SAA to this new dataset.

The choice of bandwidth parameter $h$ is a matter for careful choice. As $h$ gets small the behaviour of $\widehat{f}$ gets closer to a set of delta functions at the original data points. The result is that a solution for P$_{\text{smooth}}$ gets closer to a solution to the original problem SAA, and the two will be the same in the limit of small $h$.

\subsection{Maximum Likelihood Estimation}

The underlying idea of maximum likelihood estimation (MLE) can be applied whenever we can specify a set of possible distributions from which the observed sample might have been drawn. A maximum likelihood estimator simply chooses that distribution which gives the highest likelihood of obtaining the sample that has been observed. For example if we make no restriction at all on the underlying distribution then the maximum likelihood is obtained by choosing the (discrete) sample distribution which has probability $1/N$ at each of the points $y_{i}$, $i=1,2,...,N$. And hence we recapture the SAA solution. 

In practice we use log likelihood, and hence we choose a density function $f$ to maximize the log likelihood of observing the sample $S$, i.e. $\Sigma_{i=1}^{N} \log f(y_{i} ) $. This MLE approach can be applied for almost any set of restrictions placed on the underlying distribution. For example we have experimented with a maximum likelihood estimator for a univariate distribution with the restriction that there is a single mode together with bounds on the slope of the density function. The log likelihood calculation for any proposed density function depends only on the $f$ values at the sample points, so for a general set of distributions we only need to convert the restriction on the distribution to a restriction on these $f$ values and then treat this as a finite dimensional optimization problem. However, even though the computations are straightforward, the results in practice are very disappointing and we will not report them here. 

The most common form of maximum likelihood estimator occurs when the decision maker knows that the distribution comes from a parameterized family of distributions, for example the sample might be drawn from a normal distribution with unknown mean and variance. MLE is then used to estimate the parameters of the unknown underlying distribution. More formally we can describe this by supposing that we have a family of distributions defined by density functions $f(y;\theta)$ where the density is determined by the choice of parameters (a vector) $\theta\in\Theta$. Then we select the value of $\theta$ that maximizes the log likelihood of observing the sample. Hence we choose $\theta^{\ast}$ that solves
\[
\max_{\theta}\sum_{i=1}^{N}\log(f(y_{i};\theta))
\]
We then proceed to solve the original problem using the estimated distribution with density $\widehat{f}(y)=f(y;\theta^{\ast})$. As usual in our computations we take a large number of samples from $\widehat{f}$ and then use SAA to find the maximum likelihood choice for $x$.

\subsection{Bayesian approach}

A Bayesian framework avoids the selection of a single distribution as an estimate of the true underlying distribution. It is most useful when there is a parameterized family of distribution densities $f(y;\theta)$, $\theta\in\Theta$ from which the sample may have been drawn. However we do not need to put all of these possible distributions on an equal footing, instead we may specify what we think is the probability attached to different possible values of the parameters $\theta$, through giving a prior distribution $\gamma(\theta)$ defined on the set of possible parameters $\Theta$. After observing the sample $S=\{y_{1},y_{2},....y_{N}\}$ we can update our prior beliefs about the likelihood of different parameter values to obtain a posterior distribution $\gamma(\theta~|~S)$, where
\[
\gamma(\theta~|~S)\propto{\textstyle\prod\limits_{i=1}^{N}} f(y_{i};\theta)\gamma(\theta)
\]
and $\gamma$ is normalized to integrate to $1$. A Bayesian estimation method could select a value of the unknown parameter $\theta$ from this posterior density. However in our framework it is best to continue with an explicit uncertainty on the parameter $\theta$ and allow new sample points to be generated according to the posterior distribution $\gamma(\theta~|~S)$. The final density over which the cost function $c(x,y)$ is minimized is given by 
\[
\widehat{f_{S}}(y)=\frac{\int_{\Theta}f(y;\theta){\textstyle\prod\limits_{i=1}^{N}}f(y_{i};\theta)\gamma(\theta)d\theta}
{\int_{\Theta} {\textstyle\prod\limits_{i=1}^{N}}f(y_{i};\theta)\gamma(\theta)d\theta}
\]
This can be seen to be a density (i.e. it integrates to $1$) through interchanging the order of integration for $\theta$ and $y$. To estimate the posterior distribution MCMC sampling has been used with specific priors described later. 

\section{Computational Testing}

We will carry out computational testing on two different types of problem, first some small scale cases where both the decision variable $x$ and the random variable $Y$ are scalars and the cost function is quadratic, and second a simple portfolio optimization problem where $x$ and $y$ are multivariate. At the same time as specifying the structure of the problem we need to specify the actual underlying distribution. This will be unknown to the decision maker, but will enable us to see how good the decisions are when tested out-of-sample. Before giving more details on the two problem settings we describe the process we use for testing different methods.

The assessment of how well a method works in each case follows the same scheme. First we choose a ``true'' underlying distribution. Then we take a number, $K$, of samples each of size $N$, call these $S_{1},S_{2},...,S_{K}$ and apply the chosen approach to each sample. This generates for each sample an optimal solution call it $x_{M}(S_{i})$ which we can compare against the sample average solution $x_{SAA}(S_{i})$ (both of which depend on the sample chosen). Then we compare performance on average (across the $K$ different samples) when the expectations are taken with respect to the true underlying distribution. This final step evaluates the improvement against SAA:
\begin{equation}
\frac{1}{K}\sum_{j=1}^{K}\mathbb{E}[c(x_{SAA}(S_{j}),y)]-\frac{1}{K}\sum_{j=1}^{K}\mathbb{E}[c(x_{M}(S_{j}),y)]. \label{eq:improvement}
\end{equation}
The expectations here could potentially be done exactly through evaluating an integral with respect to the true distribution, but for our purposes it is sufficient to simply take another much larger sample from the true distribution and evaluate the average cost on that sample. At this point it will be preferable to take matched pair of samples for the two expectations in (\ref{eq:improvement}). In other words we take a large sample $\{y_{1}^{(a)},y_{2}^{(a)},...y_{L}^{(a)}\}$ and then estimate the improvement from
\[
\frac{1}{K}\sum_{j=1}^{K}\frac{1}{L}\sum_{i=1}^{L}\left(  c(x_{SAA}(S_{j}),y_{i}^{(a)})-c(x_{M}(S_{j}),y_{i}^{(a)})\right)  .
\]

\subsection{Quadratic test problems}

\begin{figure}
    \centering
    \includegraphics[width=\linewidth]{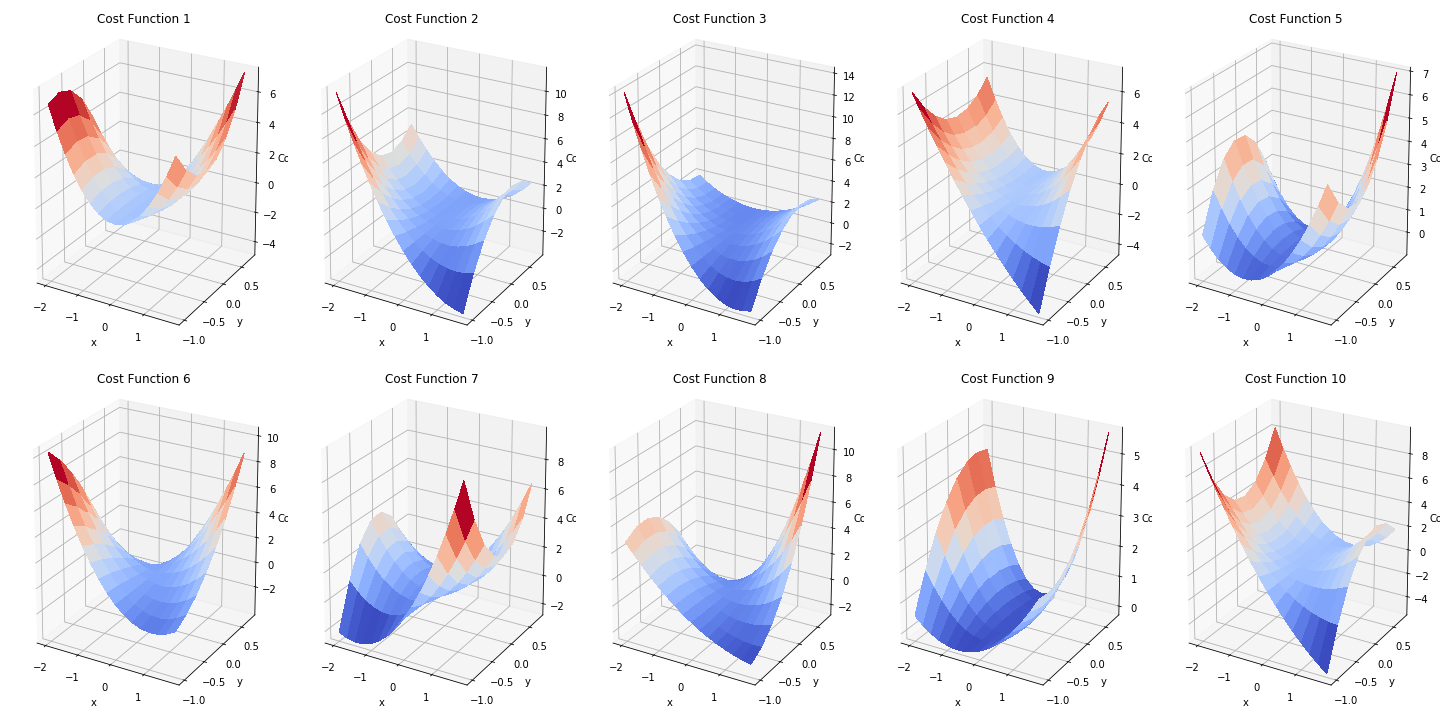}
    \caption{A set of quadratic cost functions of the form $x^2+\alpha x^2y+\beta xy^2 + \gamma xy$, where $\alpha$, $\beta$ and $\gamma$ are listed in Table \ref{tab:cost_coefs}.}
    \label{fig:cost_functions}
\end{figure}

We will use a set of 10 test problems with scalar values for the decision variable $x$ and scalar random variable $Y$, and a cost function which has at most quadratic terms in $x$ and $y$ of the form $x^2+\alpha x^2 y + \beta x y^2+ \gamma x y $ where the values $\alpha$, $\beta$, $\gamma$ for the 10 problems are given in Table \ref{tab:cost_coefs}. These sets of coefficients were selected from a larger randomly generated set in order to exhibit different types of behavior. The cost function surfaces in terms of the decision variable $x$ and the random variable $y$ for these 10 test problems are shown in Figure \ref{fig:cost_functions}. 

\begin{figure}
\begin{floatrow}
\capbtabbox{%
  \begin{tabular}{|c|c|c|}
\hline
  ID& Distribution & Parameters   \\
     \hline
  1  & Beta & \begin{tabular}{@{}c@{}}$\alpha=2$ \\ $\beta=2$\end{tabular}\\
  \hline
2  & Beta  & \begin{tabular}{@{}c@{}}$\alpha=5$ \\ $\beta=5$\end{tabular}  \\
\hline
3  & Beta & \begin{tabular}{@{}c@{}}$\alpha=2$ \\ $\beta=5$\end{tabular}   \\
\hline
4  & Gaussian mixture &	\begin{tabular}{@{}cc@{}}
                        $\mu_1=-0.5$&$\sigma_1=0.15$ \\         
                        $\mu_2=0.4$&$\sigma_2=0.3$\\
                        $\rho=0.6$
                        \end{tabular} \\
\hline
5  & Gaussian mixture &	\begin{tabular}{@{}cc@{}}
                        $\mu_1=-0.1$&$\sigma_1=0.3$ \\         
                        $\mu_2=0.4$&$\sigma_2=0.1$\\
                        $\rho=0.7$
                        \end{tabular} \\
\hline

\end{tabular}
}{%
  \caption{Parameter values for different distributions.}
 \label{tab:dist_params}
}
\ffigbox{%
  \includegraphics[width=\linewidth]{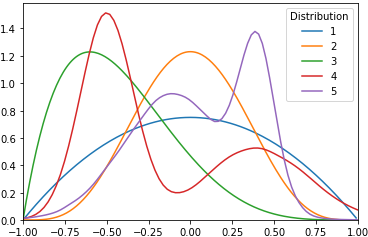}%
}{%
  \caption{Probability density function of various distributions with parameters given in Table \ref{tab:dist_params}.}
    \label{fig:distributions}
}
\end{floatrow}
\end{figure}

Each cost function is evaluated against 5 underlying ``true''  univariate distributions: three beta distributions, with parameters $\alpha$ and $\beta$ with a support of $[-1,1]$ and two bi-modal Gaussian mixture, with parameters $\mu$ and $\sigma$ for each Gaussian, and $p$, the mixing proportion of the two Gaussians. The parameter values are shown in Table \ref{tab:dist_params}. A set of $N$ samples, where $N=\{10,20,50\}$, was sampled from the given distribution and then the decision variable, $x$, was computed for each quadratic test problem. This was repeated 1000 times. The decision variable was calculated using bagging, maximum likelihood estimation, kernel smoothing, a Bayesian framework and sample average approximation. 

\begin{table}
\centering
\begin{tabular}{|c|c|c|c|}
\hline
  Cost function & $\alpha$ &$\beta$ &$\gamma$   \\
     \hline
  1  & -0.67 & 2.56  & 2.51   \\
2  & 0.02  & -2.57 & 1.31   \\
3  & -0.51 & -2.1  & 1.97   \\
4  & 0.71 &	-1.45	&1.62   \\
5  & 0.19  & 2.17  & 1.04   \\
6  & -0.26 & 1.23  & 3.89   \\
7 & -0.22 & 3.71  & 0.19 \\
8 & 0.65 &	2.02	&3.68 \\
9 & 0.6 &	0.86	&0.33 \\
10 & 0.49&	-3.25&	0.65 \\
\hline

\end{tabular}
\caption{Value of coefficients for cost functions.}
 \label{tab:cost_coefs}
\end{table}

In bagging, each new sample set was the same size as the original samples, $M=N$, and this was repeated $B=400$ times. The kernel smoothing method used a Gaussian kernel where the bandwidth parameter was chosen through Scott's Rule \cite{scott2015multivariate} and hence changes with the sample size. The MLE method when applied to the first three cases was constrained to fit a uni-modal beta distribution on $[-1,1]$ with 2 (shape) parameters, $\alpha>1$ and $\beta>1$. For the mixture of Gaussian distributions all 5 parameters were estimated using the parameter ranges of $\sigma_i > 0.1$.  The Bayesian framework for the first three distributions used uniform priors with a support of $[1,7]$ for $\alpha$ and $\beta$. For distributions 4 and 5, to estimate model parameters $\mathbf{\mu}$, $\mathbf{\sigma}^2$ and $\mathbf{\rho}$ we used conjugate priors as described by \cite{bishop2006pattern}:
\begin{equation}
\begin{split}
    \rho_k &\sim Dir(\delta) \\
  \frac{1}{\sigma_k^2} & \sim Wishart(V,n) \\
  \mu_k | \sigma_k & \sim N(m,\sigma_k^2/\alpha) \quad for \quad k=1,2\\
\end{split}
\end{equation}
where $\alpha=0.1$, $V=0.1$, $n=2$, $\delta=10$ and $m$ set to the mean of samples. 
\begin{figure*}[t!]
    \centering
    \begin{subfigure}[t]{0.45\textwidth}
        \centering
        \includegraphics[width=\textwidth]{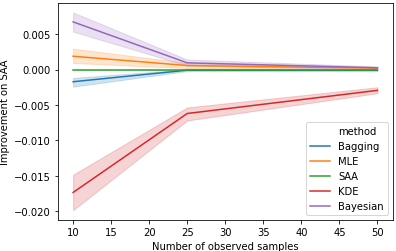}
        \caption{Distribution 1}
    \end{subfigure}%
    ~ 
    \begin{subfigure}[t]{0.45\textwidth}
        \centering
        \includegraphics[width=\textwidth]{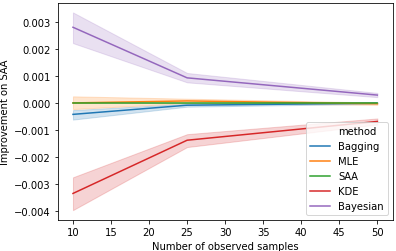}
        \caption{Distribution 2}
    \end{subfigure}
    ~
    \begin{subfigure}[t]{0.45\textwidth}
        \centering
        \includegraphics[width=\textwidth]{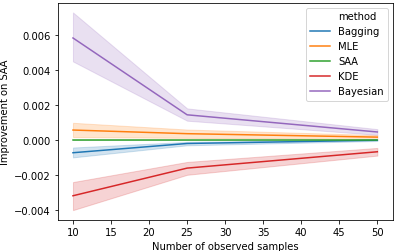}
        \caption{Distribution 3}
    \end{subfigure}
    ~
    \begin{subfigure}[t]{0.45\textwidth}
        \centering
        \includegraphics[width=\textwidth]{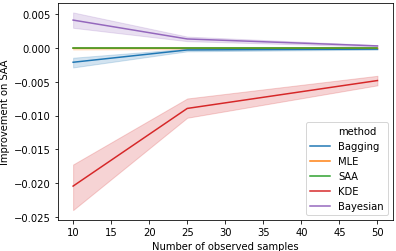}
        \caption{Distribution 4}
    \end{subfigure}
    ~
    \begin{subfigure}[t]{0.45\textwidth}
        \centering
        \includegraphics[width=\textwidth]{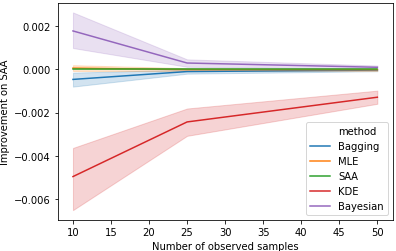}
        \caption{Distribution 5}
    \end{subfigure}
    
    \caption{The improvement over SAA averaged over cost functions for the set of underlying distributions  with shading for 95\% confidence interval.}
    \label{fig:results-quadratic-plots}
\end{figure*}

Figure \ref{fig:results-quadratic-plots} shows the results we obtain on this test set, which we will discuss in more detail in the next section.

\subsection{Portfolio optimization test problems}

Portfolio optimization problems are often used to test stochastic optimization approaches. They are of interest in practice and provide a significant challenge to obtain good out of sample performance. Moreover it is known that the SAA approach can perform poorly on these problems.

There is a wealth of empirical data available for stock returns, however we have chosen to carry out our tests on synthetic data. The reason for this choice is that for both MLE and Bayesian approaches we need to have a parametric model of the underlying process. In the case of stock prices there are many potential models and the performance on empirical data of the methods we investigate will critically depend on the accuracy of the model chosen. This will have the effect of making it harder to assess the value of different approaches. By using synthetic data we can avoid these problems.

Given a set of $n$ stocks, the most common form of portfolio optimization problem seeks weights $w_{i}$, $i=1,2,...,n$ (these are the decision variables we have normally labelled $x$) to minimize risk subject to a target for portfolio return. If we measure risk by the variance of the returns, as in the standard Markowitz version of portfolio theory, we get the problem 
\[
\begin{tabular}{ll}
$\min_{w}$ & $\mathbb{E}\left[ w^{\top }\Sigma w\right] $ \\ 
subject to & $w^{\top }1_{n}=1,$ \\ 
& $w^{\top }\mu =R.$ 
\end{tabular}
\ \ 
\]
where $\Sigma $ is the covariance matrix for the stock returns, $\mu $ is the vector of mean returns and $R$ is a target portfolio return. In this version of the problem we allow borrowing (corresponding to negative components in $w$). 

Given a sample over $N$ periods of stock returns we can replace $\Sigma $ by $\widehat{\Sigma }$ which is the sample covariance, and replace $\mu $ by the sample mean, $\widehat{\mu }$. Solving this in-sample problem then gives a set of portfolio weights that can be assessed out of sample. 

In practice the target portfolio return $R$ is hard to determine. Actual market returns can vary significantly from period to period and so the target becomes somewhat arbitrary. Moreover the estimates of mean return for individual stocks are important for the solution but are typically unreliable. Hence it is quite common to drop the target return constraint and simply consider a minimum variance portfolio given a set of stocks that are all expected to have reasonably good mean returns. This minimum variance portfolio problem is the one that we will consider. Thus with $W=\{w:\sum_{i=1}^{n}w_{i}=1\}$ we solve $\min_{w\in W}[w^{\top }\widehat{\Sigma }w]$. 

Let $z_{i}^{(k)}$ be the return for stock $i$ in period $k$ (i.e. the $k$'th element of the sample). We have $\widehat{\Sigma }_{ij}=\frac{1}{N} \sum_{k=1}^{N}(z_{i}^{(k)}-\overline{z}_{i})(z_{j}^{(k)}-\overline{z}_{j})$ where $\overline{z}_{i}=\frac{1}{N}\sum_{k=1}^{N}z_{i}^{(k)}$, $i=1,2,...n$, (noting that whether we use $1/N$ or $1/(N-1)$ in $\widehat{\Sigma }$ will not change the optimal weights). Thus  
\begin{eqnarray*}
w^{\top }\widehat{\Sigma }w &=&\sum_{i}\sum_{j}w_{i}w_{j}\frac{1}{N}\sum_{k=1}^{N}(z_{i}^{(k)}-\overline{z}_{i})(z_{j}^{(k)}-\overline{z}_{j}) \\
&=&\frac{1}{N}\sum_{k=1}^{N}\sum_{i}\sum_{j}w_{i}w_{j}z_{i}^{(k)}z_{j}^{(k)}-\left( \sum_{i}w_{i}\overline{z}_{i}\right) \left( \sum_{j}w_{j}\overline{z}_{j}\right)  
\end{eqnarray*}

In general the in-sample minimum variance portfolio problem cannot be expressed as simply an SAA solution with objective formed as the sum of costs over each of the samples $z^{(k)}$. However, in the case that the mean returns from each stock are the same with $\overline{z}_{i}=\overline{z}$, then the second term in the expression above becomes $\overline{z}^{2}$. In this case we get an optimal solution for the in-sample problem which matches the SAA solution with cost function $c(w,z)=\left( w^{\top }z\right)^{2}$. 

In our experiments we use 5 different covariance matrices and assume a t-distribution for individual stock returns with 3 degrees of freedom each with 0 mean. This is a good fit for real data \cite{peiro1994distribution}. The covariances for these ``true'' distributions are estimated from real stock data (using S\&P 500 data for weekly returns). 

For the MLE approach, the mean and covariance were estimated assuming 3 degrees of freedom.  The Bayesian framework used the following model: \begin{equation} \label{eqn:portfolio_bayesian_model}
\begin{split}
    \sigma &\sim Gamma(\alpha,\beta) \\
  \bm{L} | \sigma &\sim LKJCov(\eta,n,\sigma)\\
  \bm{\mu} | \bm{L} &\sim \mathcal{N}(\mathbf{0},\mathbf{L}\mathbf{L}^T)\\
  \bm{s_i} | \bm{\mu},\bm{L} &\sim t(\bm{\mu},\mathbf{L}\mathbf{L}^T,\nu) \quad for \quad i=1,...,N,\\
\end{split}
\end{equation}
where $\alpha=3$, $\beta=1$, $\eta=2$, $n=5$ and $\nu=3$. The LKJCov distribution is a distribution over cholesky decomposed covariance matrices, such that the underlying correlation matrices follow an LKJ distribution. This allows for more efficient sampling of covariance matrices \cite{lewandowski2009generating}.

\begin{figure*}[t!]
    \centering
    
    \begin{subfigure}[t]{0.45\textwidth}
        \centering
        \includegraphics[width=\textwidth]{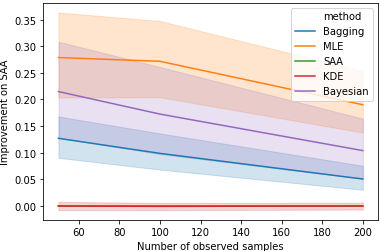}
        \caption{Covariance 1}
    \end{subfigure}%
    ~ 
    \begin{subfigure}[t]{0.45\textwidth}
        \centering
        \includegraphics[width=\textwidth]{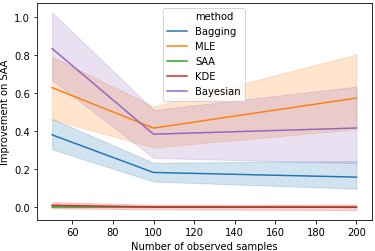}
        \caption{Covariance 2}
    \end{subfigure}
    ~
    \begin{subfigure}[t]{0.45\textwidth}
        \centering
        \includegraphics[width=\textwidth]{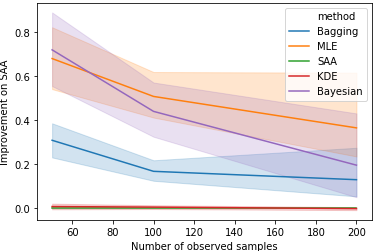}
        \caption{Covariance 3}
    \end{subfigure}
    ~
    \begin{subfigure}[t]{0.45\textwidth}
        \centering
        \includegraphics[width=\textwidth]{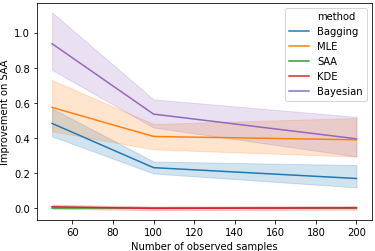}
        \caption{Covariance 4}
    \end{subfigure}
    ~
    \begin{subfigure}[t]{0.45\textwidth}
        \centering
        \includegraphics[width=\textwidth]{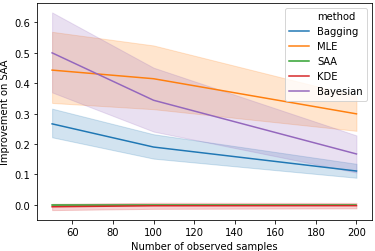}
        \caption{Covariance 5}
    \end{subfigure}
    \caption{The improvement over SAA optimising a portfolio for a set of 5 different covariances with shading for 95\% confidence interval.}
    \label{fig:results-portfolio}
 \end{figure*}
 
 Figure \ref{fig:results-portfolio} shows the results we obtain for the 5 different portfolio optimization problems considered.
 
\section{Discussion and Conclusion}

We begin by discussing the results for the quadratic test set shown in Figure \ref{fig:results-quadratic-plots}. As we would expect the greatest differences in performance occur with small sample sizes. First note that kernel density smoothing performs poorly in all cases. Our test cases involve cost functions that are already smooth with respect to changes in the random variable $y$, and so there is nothing to be gained by smoothing the density functions. More surprisingly we find a significant degradation in performance. 

Also in these experiments there is no advantage gained from bagging, with a small performance penalty for sample size 10.  Moreover the maximum likelihood estimator makes only a small improvement on SAA, with no improvement at all on 3 of the 5 distributions. This is a surprise and we have carried out some additional experiments to explore what is happening. In this case MLE performance can be significantly weakened by over-fitting.  The results presented here have only a small number of estimated parameters. For the beta distributions we assume that the end points of the support of the distribution are known. If these parameters are unknown and also fitted using the MLE approach, then the end results are significantly worse than for SAA.  

The clear winner for these quadratic test problems is a Bayesian approach, which is able to take advantage of the information on the underlying distributions without problems of  over fitting associated with MLE. For example with beta distributions when we need to estimate the end points of the distribution support, a Bayesian approach is still effective, even though MLE does worse than SAA in this case due to over fitting. 

When we consider the portfolio optimization problems we obtain rather different results. First observe that kernel smoothing has essentially the same performance as SAA. It is interesting that with portfolio optimization bagging performs well, with an improvement on SAA in all cases. This method makes no use of information on the underlying distribution but achieves between 30\% and 40\% of the available benefits of the more sophisticated approaches.

MLE and Bayesian approaches both do well, with the Bayesian approach being best with a sample size of 50 in 4 out of 5 cases, but the MLE is preferred for larger sample sizes. 

In summary, for one dimensional problems it is hard to beat SAA, but where there is information on a parameterized family of distributions then a Bayesian estimation is recommended. Maximum Likelihood estimation should not be used. For multi-dimensional problems, like portfolio optimization, bagging is a technique that should be applied in the absence of good information on the underlying distribution. Where there is a parameterized family of distributions then either MLE or Bayesian methods can work well.

\bibliography{SAArefs}

\end{document}